\documentclass[conference]{IEEEtran}
\IEEEoverridecommandlockouts
\usepackage{cite}
\usepackage{amsmath,amssymb,amsfonts}
\usepackage{graphicx}
\usepackage{textcomp}
\usepackage{xcolor}
\usepackage[utf8]{inputenc} % allow utf-8 input
\usepackage[T1]{fontenc}    % use 8-bit T1 fonts
\usepackage{hyperref}       % hyperlinks
\usepackage{url}            % simple URL typesetting
\usepackage{booktabs}       % professional-quality tables
\usepackage{amsfonts}       % blackboard math symbols
\usepackage{nicefrac}       % compact symbols for 1/2, etc.
\usepackage{microtype}      % microtypography
\usepackage{lmodern}
\usepackage[noend]{algpseudocode}
\usepackage{algorithmicx}
\usepackage{algorithm}
\usepackage{algpseudocode}

\def\BibTeX{{\rm B\kern-.05em{\sc i\kern-.025em b}\kern-.08em
    T\kern-.1667em\lower.7ex\hbox{E}\kern-.125emX}}
\begin{document}

\title{Cognitive Consistency Routing Algorithm of Capsule-network\\}

\author{\IEEEauthorblockN{1\textsuperscript{st} Huayu Li}
\IEEEauthorblockA{\textit{Northern Arizona University}\\
Flagstaff, United State\\
\textit{Northern Arizona University}\\
hl459@nau.edu}
}

\maketitle

\begin{abstract}
Artificial Neural Networks (ANNs) are computational models inspired by the central nervous system (especially the brain) of animals and are used to estimate or generate unknown approximation functions that rely on a large amount of inputs. The Capsule Neural Network \cite{bibitem1} is a novel structure of Convolutional Neural Networks(CNN) which simulates the visual processing system of human brain. In this paper, we introduce a psychological theory which is called Cognitive Consistency to optimize the routing algorithm of Capsnet to make it more close to the working pattern of human brain. Our experiments show that progress had been made compared with the baseline.
\end{abstract}


\section{Introduction}
Convolutional neural networks(CNN) contribute to a series of breakthroughs for image classification task. There are plenty of structures of CNN proven to make an outstanding performance in classification tasks in various domains. Capsnet is a novel variant of CNN proposed by \cite{bibitem1}. Capsnet uses the outputs of a group of neurons which is called capsule\cite{bibitem2} to represent different properties of the same entity. The mechanism\cite{bibitem1} of Capsnet is to ensure that the output of the capsule gets sent to an appropriate parent in the layer above. Initially, the output is routed to all possible parents after scaled down by coupling coefficients that sum to 1. A “prediction vector” which is the product of its own output and a weight matrix for each possible parent is computed by the capsule. Top-down feedback increases the coupling coefficient for that parent and decreases coupling coefficient for other parents, determined by the scalar product of this prediction vector and the output of a possible parent. The parent further increases the scalar product of the capsule’s prediction with the parent’s output as a result of increasing the contribution made by the capsule. In \cite{bibitem1}, the “agreement” is mentioned as simply the scalar product $a_{j}=v_{j}\cdot\hat{u}_{j|i}$ which is treated as if it was a log likelihood and is added to the initial logit, $b_{ij}$ before computing the new values for all the coupling coefficients linking capsule $i$ to higher level capsules.

In this paper, we were inspired by psychological theories and applied them to develop a new method of routing algorithm. According to \cite{bibitem3}, people have a drive to generate consistent cognition and behavior on objects. When the cognition is dissonant, people will feel uncomfortable, and then try to reduce it, reducing a mechanism of dissonance by selectively seeking support information or avoiding inconsistent information.

We simply regard the prediction vectors made by each capsule as the “cognition” it keeps, and our work is to explore an algorithm to unify the cognition between each capsule, which we call the Cognitive Consistency Routing Algorithm. We propose to treat the differences between the input “prediction vectors” from the lower layer and the “prediction vectors” made by the current layer as dissonance coefficients to ensure the coupling between higher capsules and lower layer increases much sooner than if the differences between their predictions are smaller. The motivation behind the Cognitive Consistency Routing Algorithm is to ensure each capsule layer makes the prediction of the target is as consistent as possible. In other words, the capsule layers of the whole network should ultimately tend to a consistent cognition.

\section{Cognitive Consistency Routing Algorithm}
There are many possible ways to implement the general idea of Cognitive Consistency. The aim of this paper is not to find a method that achieves the state-of-art performance but simply to test and verify the practicability of the Cognitive Consistency Routing Algorithm.

Firstly, we want to get the initial values of the prior probabilities from the lower capsules and let the capsules discriminatively coupled to the lower layers. We therefore use a simple “clip” function for the prior probabilities initially to ensure that the capsules above accept the prediction vectors $\hat{\mathbf{u}}_{j|i}$ from the capsules below and restrict them within a range.
\begin{equation}
  b_{ij} =
  \begin{cases}
    a_{min}, &\text{if $\hat{\mathbf{u}}_{j|i}<a_{min}$}\\
        \hat{u}_{j|i}, &\text{if $a_{min}<\hat{\mathbf{u}}_{j|i}<a_{max}$}\\
	       a_{max}, &\text{if $\hat{\mathbf{u}}_{j|i}>a_{min}$,}
  \end{cases}
\end{equation}
where the $\hat{\mathbf{u}}_{j|i}$ are the prediction vectors from the lower capsules and $(a_{min},a_{max})$ is the expected range of $b_{ij}$. The motivation behind “clip” function is to avoid completely inactivating the prediction vectors. We expect the "bad guys" which are decided by the lower capsule continue to have the right to make predictions in current layer thus the whole network maintain a consistent cognition without partial dissonance.

The coupling coefficients $c_{ij}$ between capsule $i$ and all the capsules in the layer above sum to 1 are determined by the distribution of the prior probabilities by the “softmax” function.
\begin{equation}
    c_{ij}=\frac{exp(b_{ij})}{\sum _{k}exp(b_{ik})}
\end{equation}

Our approach to get the total input to a capsule $s_{j}$ is as the same as Sabour S, et al.[2017].
\begin{equation}
    \mathbf{s}_{j}=\sum_{j}c_{ij}\hat{\mathbf{u}}_{j|i}, \hat{\mathbf{u}}_{j|i}=\mathbf{W}_{ij}\mathbf{u}_{i}
\end{equation}
where $\mathbf{u}_{i}$ is the output of capsule in the layer below and $\mathbf{W}_{ij}$ is the weight matrix.

And the vector output of capsule $j$ is the “squashing” results of $\mathbf{s}_{j}$.
\begin{equation}
    \mathbf{v}_{j}=\frac{\left \| \mathbf{s}_{j} \right \|^{2}}{1+\left \| \mathbf{s}_{j} \right \|^{2}}\frac{\mathbf{s}_{j}}{\left \| \mathbf{s}_{j} \right \|}
\end{equation}
which ensure that the vectors are in the range of zero to one.

The prior probabilities  are supposed to be iterated by adding the scalar product $\left \| \mathbf{v}_{j} \right \|\cdot \left \| \hat{\mathbf{u}}_{j|i} \right \|\cdot a_{ij}$.
\begin{equation}
b_{ij}=b_{ij}+\left | \mathbf{v}_{j} \right |\cdot \left | \hat{\mathbf{u}}_{j|i} \right |\cdot a_{ij},a_{ij}=cos((\left |\mathbf{v}_{j}\right |-\left |\hat{\mathbf{u}}_{j|i}\right |)^{2})
\end{equation}
where $a_{ij}$ are the consistency ratios that decreases with the increment of difference between the input prediction vectors from the lower layer and the prediction vectors made by the current layer. The curve of $a_{ij}$ is shown as Fig. 1.
\begin{figure}[htbp]
	\centering
    \includegraphics[scale=0.4]{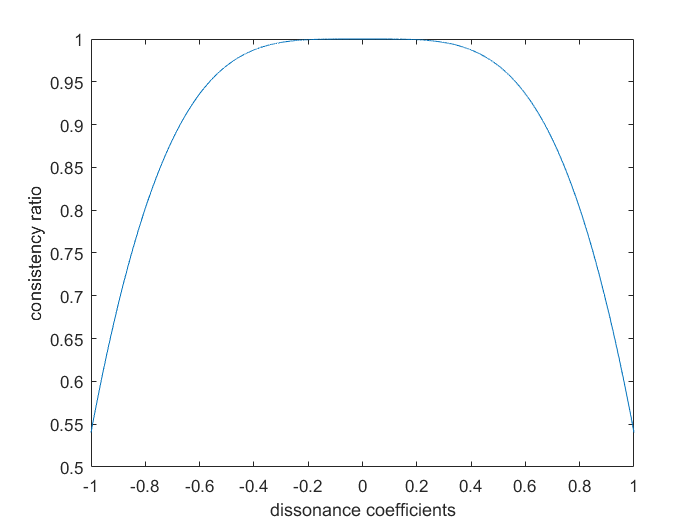}
	\caption{The curve of consistency ratios shows that the bigger difference between the predictions made by the current layer and the lower layer lead to the smaller consistency ratios which means each layer tends to a consistent cognition to avoid cognitive dissonance.}
	\label{fig:lable}
\end{figure}

The overall flow of our algorithm is shown in Algorithm 1.

\begin{algorithm}[htb]
  \caption{Cognitive Consistency Routing Algorithm of Capsule-network}
  \begin{algorithmic}[1]
    \State For all capsule $i$ in layer $l$ and capsule $j$ in layer $(l+1)$: $b_{ij}\leftarrow clip(\hat{\mathbf{u}}_{j|i},a_{min},a_{max})$ .
    \For{each $r$ iterations}
      \State for all capsule $i$ in layer $l$: $c_{ij}\leftarrow \frac{exp(b_{ij})}{\sum _{k}exp(b_{ik})};$
      \State for all capsule $j$ in layer $l+1$: $\mathbf{s}_{j}\leftarrow \sum_{j}c_{ij}\hat{\mathbf{u}}_{j|i};$
      \State for all capsule $j$ in layer $l+1$: $\mathbf{v}_{j}\leftarrow squash(\mathbf{s}_{j});$
      \State for all capsule $i$ in layer $l$ and all capsule $j$ in layer $l+1$: $b_{ij}\leftarrow b_{ij}+\left \| \mathbf{v}_{j} \right \|\cdot \left \| \hat{\mathbf{u}}_{j|i} \right \|\cdot cos((\left |\mathbf{v}_{j}\right |-\left |\hat{\mathbf{u}}_{j|i}\right |)^{2});$
    \EndFor
    \\
    \Return $\mathbf{v}_{j};$
  \end{algorithmic}
\end{algorithm}

\section{Experimental Results}
\subsection{Our Algorithm on MNIST}
We tested our algorithm on MNIST to verify whether our algorithm works or not. Our model has the same architecture as \cite{bibitem1} showed in Fig. 2 and was set as 3 times routing, but with batch normalization(BN)\cite{bibitem4} after each layer. We tested two algorithms on this model and compared their performance. We observed that a Capsnet with Cognitive Consistency Routing achieves state-of-the-art performance on MNIST which equals to the baseline.

\begin{figure}[htbp]
	\centering
    \includegraphics[scale=0.4]{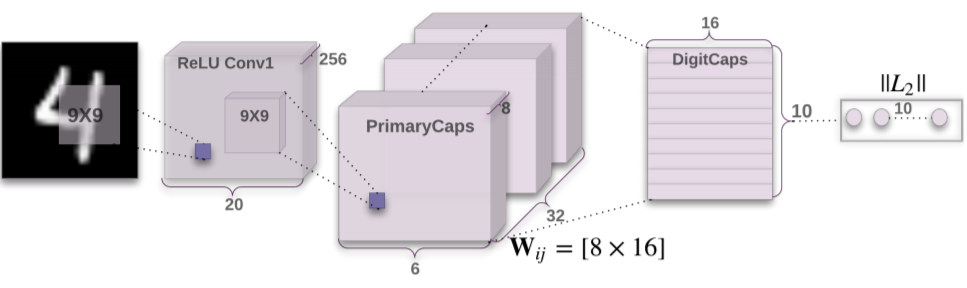}
	\caption{The structure of the original Capsnet.}
	\label{fig:lable}
\end{figure}

\subsection{Our Algorithm on Fashion-MNIST}
Via experiments on other datasets, we can simply observe that our algorithm get better and higher results than the baseline in more complex data. We use the same model as on MNIST. We evaluated our method and the original algorithm on the Fashion-MNIST dataset \cite{bibitem5} which consists of 28x28 grayscale images of 70,000 fashion products from 10 categories, with 7,000 images per category. The models were trained on the 48,000 training images and evaluated on the 12,000 validation images. We also obtained a final result on the 10,000 test images. We tested using 10 models and got the final result by model averaging. We can see from table 1 that our routing approach achieved better performance than the baseline.

\begin{table}[htbp]
\centering
\begin{tabular}{ccccccc}
\hline
Iterations& 1& 3& 5& 10 & 15\\
\hline
Sabour S, et al.[2017] & 0.7632& 0.7993& 0.8227& 0.8514& 0.8644\\
Our Algorithm& 0.7832& 0.8163& 0.8266& 0.8655& 0.8803\\
\hline
\end{tabular}
\caption{Our Algorithm on Fashion-MNIST, the higher the better as the original Algorithm}
\end{table}

\section{Conclusion}
We show that the Cognitive Consistency from psychological theories can improve the capsule neural network and demonstrate that introducing the theories from other domains, besides only statistics or computer science is an effective way to perfect the theoretical foundation of deep learning and Artificial Neural Networks. Future work includes improving our algorithm to achieve the state-of-art performance and trying to introduce more theories from psychologic and cognitive science into Artificial Neural Networks.

\section*{Acknowledgment}
At the point of finishing this paper, we would like to express our sincere thanks to the authors of Dynamic routing between capsules\cite{bibitem1} who had made great breakthroughs and outstanding contributions in exploring new architecture of neural networks.
At the same time, we also have to thank Professor.Razi and Professor.Bakke who gave guidance and support to us during the process of completing this paper.


\begin{thebibliography}{3}
\bibitem{bibitem1}Sabour S, Frosst N, Hinton G E. Dynamic routing between capsules[C] Advances in Neural Information Processing Systems. 2017: 3856-3866.
\bibitem{bibitem2}Hinton G E, Krizhevsky A, Wang S D. Transforming auto-encoders[C] International Conference on Artificial Neural Networks. Springer, Berlin, Heidelberg, 2011: 44-51.
\bibitem{bibitem3}Festinger L. A theory of cognitive dissonance[M]. Stanford university press, 1962.
\bibitem{bibitem4}Ioffe S, Szegedy C. Batch normalization: Accelerating deep network training by reducing internal covariate shift[J]. arXiv preprint arXiv:1502.03167, 2015.
\bibitem{bibitem5}Xiao H, Rasul K, Vollgraf R. Fashion-mnist: a novel image dataset for benchmarking machine learning algorithms[J]. arXiv preprint arXiv:1708.07747, 2017.
\end{thebibliography}
\end{document}